\PassOptionsToPackage{dvipsnames}{xcolor}
\documentclass[sigconf]{acmart}

\usepackage{graphicx}

\usepackage{amsmath,amssymb}   
\usepackage{booktabs}          
\usepackage{algorithm}
\usepackage{algorithmic}
\usepackage{comment}
\usepackage[inline]{enumitem}  
\definecolor{niceblue}{rgb}{0.702, 0.702, 1} 
\definecolor{nicered}{rgb}{1, 0.702, 0.702} 

\newcommand{\celsystem}{\textsc{PruneCEL}}
\newcommand{\celoe}{\textsc{CELOE}}
\newcommand{\drill}{\textsc{Drill}}
\newcommand{\dlfoil}{\textsc{DL-Foil}}
\newcommand{\evolearner}{\textsc{EvoLearner}}
\newcommand{\nces}{\textsc{NCES}}

\newcommand{\deeppavlov}{\textsc{DeepPavlov}}
\newcommand{\ganswer}{\textsc{gAnswer}}
\newcommand{\mst}{\textsc{MST5}}
\newcommand{\tebaqa}{\textsc{TeBaQA}}

\newcommand{\etal}{et~al.}
\newcommand{\doi}[1]{\href{https://doi.org/#1}{doi:#1}}

\newcommand{\surveyParticipants}{41}

\newcommand{\qasystems}{4}

\newcommand{\alc}{\ensuremath{\mathcal{ALC}}}
\newcommand{\knowledgebase}{\ensuremath{\mathcal{K}}}
\newcommand{\tbox}{\ensuremath{\mathcal{T}}}
\newcommand{\abox}{\ensuremath{\mathcal{A}}}
\newcommand{\concepts}{\ensuremath{\mathcal{C}}}
\newcommand{\namedConcepts}{\ensuremath{N_C}}
\newcommand{\namedRoles}{\ensuremath{N_R}}
\newcommand{\namedIndividuals}{\ensuremath{N_I}}
\newcommand{\interpretation}{\ensuremath{\mathcal{I}}}
\newcommand{\interpret}[1]{\ensuremath{#1^{\interpretation}}}
\newcommand{\domain}{\ensuremath{\interpret{\Delta}}}
\newcommand{\length}{\ensuremath{l}}
\newcommand{\conceptC}{\ensuremath{C}}
\newcommand{\conceptD}{\ensuremath{D}}
\newcommand{\conceptX}{\ensuremath{X}}
\newcommand{\conceptY}{\ensuremath{Y}}
\newcommand{\role}{\ensuremath{r}}

\newcommand{\referenceGraph}{\ensuremath{\mathcal{G}}}
\newcommand{\Question}{\ensuremath{Q}}
\newcommand{\answers}{\ensuremath{U}}

\newcommand{\qasystem}{\ensuremath{f}} 

\newcommand{\examples}{\ensuremath{E}}
\newcommand{\Example}{\ensuremath{e}}
\newcommand{\pexamples}{\ensuremath{E^{+}}}
\newcommand{\nexamples}{\ensuremath{E^{-}}}
\newcommand{\exampleSpace}{\ensuremath{2^{\namedIndividuals}}}
\newcommand{\subpexamples}{\ensuremath{E^{+\prime}}}

\newcommand{\refine}{\ensuremath{\rho}}
\newcommand{\trefine}{\ensuremath{\rho^{\star}}}
\newcommand{\quality}{\ensuremath{q}}

\newcommand{\template}{\ensuremath{T}}
\newcommand{\templates}{\ensuremath{\mathcal{C}^{\star}}}
\newcommand{\marker}{\ensuremath{\mu}}
\newcommand{\tmerge}{\ensuremath{m^{\star}}}
\newcommand{\cmerge}{\ensuremath{m}}

\newcommand{\generator}{\ensuremath{g}}
\newcommand{\oracle}{\ensuremath{o}} 
\newcommand{\selconcept}{\ensuremath{{\oracle}_c}}
\newcommand{\selnegconcept}{\ensuremath{{\oracle}_{\neg c}}}
\newcommand{\selrole}{\ensuremath{{\oracle}_r}}

\newcommand{\measure}{\ensuremath{h}}
\newcommand{\poscount}{\ensuremath{p}}
\newcommand{\negcount}{\ensuremath{n}}

\copyrightyear{2025}
\acmYear{2025}
\setcopyright{cc}
\setcctype{by}
\acmConference[K-CAP '25]{Knowledge Capture Conference 2025}{December 10--12, 2025}{Dayton, OH, USA}
\acmBooktitle{Knowledge Capture Conference 2025 (K-CAP '25), December 10--12, 2025, Dayton, OH, USA}
\acmDOI{10.1145/3731443.3771359}
\acmISBN{979-8-4007-1867-0/2025/12}

\begin{document}

\title{Explainable Benchmarking through the Lense of Concept Learning}

\author{Quannian Zhang}
\orcid{0009-0008-9497-3204}
\affiliation{%
  \institution{Faculty of Computer Science,
Electrical Engineering and
Mathematics    \\  Data Science Group (DICE), Heinz
Nixdorf Institute, Paderborn
University}
  \city{Paderborn}
  \state{North Rhine-Westphalia}
  \country{Germany}}
\email{quannian@mail.uni-paderborn.de}

\author{Michael Röder}
\orcid{0000-0002-8609-8277}
\affiliation{%
  \institution{Faculty of Computer Science,
Electrical Engineering and
Mathematics    \\  Data Science Group (DICE), Heinz
Nixdorf Institute, Paderborn
University}
  \city{Paderborn}
  \state{North Rhine-Westphalia}
  \country{Germany}}
\email{michael.roeder@uni-paderborn.de}

\author{Nikit Srivastava}
\orcid{0009-0004-5164-4911}
\affiliation{%
  \institution{Faculty of Computer Science,
Electrical Engineering and
Mathematics    \\  Data Science Group (DICE), Heinz
Nixdorf Institute, Paderborn
University}
  \city{Paderborn}
  \state{North Rhine-Westphalia}
  \country{Germany}}
\email{nikit.srivastava@uni-paderborn.de	}

\author{N'Dah Jean Kouagou}
\orcid{0000-0002-4217-897X}
\affiliation{%
  \institution{Faculty of Computer Science,
Electrical Engineering and
Mathematics    \\  Data Science Group (DICE), Heinz
Nixdorf Institute, Paderborn
University}
  \city{Paderborn}
  \state{North Rhine-Westphalia}
  \country{Germany}}
\email{ndah.jean.kouagou@upb.de}

\author{Axel-Cyrille Ngonga Ngomo}
\orcid{0000-0001-7112-3516}
\affiliation{%
  \institution{Faculty of Computer Science,
Electrical Engineering and
Mathematics    \\  Data Science Group (DICE), Heinz
Nixdorf Institute, Paderborn
University}
  \city{Paderborn}
  \state{North Rhine-Westphalia}
  \country{Germany}}
\email{axel.ngonga@upb.de}


\begin{abstract}
Evaluating competing systems in a comparable way, i.e., benchmarking them, is an undeniable pillar of the scientific method. However, system performance is often summarized via a small number of metrics. The analysis of the evaluation details and the derivation of insights for further development or use remains a tedious manual task with often biased results. Thus, this paper argues for a new type of benchmarking, which is dubbed explainable benchmarking. The aim of explainable benchmarking approaches is to automatically generate explanations for the performance of systems in a benchmark. We provide a first instantiation of this paradigm for knowledge-graph-based question answering systems. We compute explanations by using a novel concept learning approach developed for large knowledge graphs called \celsystem{}. Our evaluation shows that \celsystem{} 
outperforms state-of-the-art concept learners on the task of explainable benchmarking by up to 0.55 points F1 measure. A task-driven user study with \surveyParticipants{} participants shows that in 80\% of the cases, the majority of participants can accurately predict the behavior of a system based on our explanations.
Our code and data are available at \url{https://github.com/dice-group/PruneCEL/tree/K-cap2025}. 

\end{abstract}



\maketitle

\section{Introduction}
\label{sec:introduction}

Comparable benchmarks are key for the improvement of solutions across disciplines with quantifiable results. This insight has led to the development of a multitude of benchmarking frameworks and online leaderboards based thereupon in recent years. Examples include the SEALS platform for link discovery~\cite{2021oaei}, GERBIL QA for question answering~\cite{usbeck2019benchmarking}, HOBBIT for big linked data applications~\cite{roeder2020hobbit}, and HuggingFace's Open LLM Leaderboard\footnote{\url{https://huggingface.co/spaces/open-llm-leaderboard/open_llm_leaderboard\#/}}. However, benchmarks are of little use if the results they generate do not lead to actionable insights. Hence, some benchmarking frameworks provide insights into evaluation results ~\cite{usbeck2015demo,Bast2022Elevant,Waitelonis2016}, e.g., by means of correlation analyses. While these approaches give insights within their respective predefined dimensions, none of these approaches goes beyond that paradigm. 

Explanation theory~\cite{rohlfing2020explanation} suggests that actionable insights must enable the explainee\footnote{That is, the entity receiving the insights.}  to perform better at system-relevant tasks, i.e., using and developing the benchmarked system in the case of benchmarking. Hence, we argue for the need for explainable benchmarking approaches that give human-understandable insights into when a system performs well and into when it is subpar. 
We instantiate this novel paradigm with the two following contributions:

\begin{enumerate}

\item We propose an approach for generating explanations for benchmarking results. Internally, our approach builds 
a structured representation of the benchmark data and uses concept learning to generate an explanation that is able to describe cases in which the benchmarked system performs well, separating them from cases in which it performs subpar.
We evaluate our approach with benchmark results from the area of knowledge-graph-based question answering (QA).
\item We propose a new concept learning algorithm dubbed \celsystem{} that achieves scalability by pruning the space in which it searches for concepts.
Our evaluation shows that \celsystem{} is able to outperform several state-of-the-art concept learning approaches on large knowledge bases.
\end{enumerate}

\section{Related-work}
\label{sec:related-work}

\subsection{QA Benchmarking}

Over the past decade, numerous QA benchmarks have been introduced, e.g., by the benchmarking series QALD~\cite{qald10}, LcQUAD~\cite{LcQUAD2}, and RuBQ~\cite{rubq}.\footnote{\url{https://qald.aksw.org/}, \url{https://github.com/AskNowQA/LC-QuAD}, and \url{https://github.com/vladislavneon/RuBQ}.}
These benchmarks either come with evaluation scripts or can be used in combination with benchmarking platforms like GERBIL QA~\cite{usbeck2019benchmarking}.
However, similar to other research areas, such evaluations typically provide a summary of key performance indicators (KPIs), e.g., the F1 measure of different evaluated QA systems on each dataset.
There are some attempts to give deeper insights into reasons why systems might perform good or subpar during an evaluation. For example, if provided with additional data GERBIL QA analyzes the system's performance in preprocessing steps of a typical QA pipeline, e.g., the identification of named entities or properties in the given question~\cite{usbeck2019benchmarking}. 
While these tools can help run evaluations and analyze the results, their analysis is bound to pre-defined features of the single tasks a system has to fulfill. 
To the best of our knowledge, we are the first to propose a generic approach for generating explanations, which avoids this limitation by transforming information available about the benchmarking process into a structured, generic representation and applying concept learning, i.e., symbolic machine learning, to learn an explanation for the benchmarking results.

While existing approaches like QED~\cite{QED2021} utilize symbolic machine learning for post-hoc explainability in QA, they are limited to explaining individual answer choices. Our work presents a novel approach that instead explains a system’s overall evaluation results on a dataset. By transforming
benchmarking information into a structured representation and applying concept learning, we move beyond explaining what a system answers to explaining where its strengths and blindspots lie.

\subsection{Concept Learning}

The application of concept learning---i.e., the task of describing a set of positive examples, separating it from a set of negative examples based on a knowledge base using description logics---as part of our approach led to the development of a new concept learning algorithm called \celsystem{}. Previous works use inductive logic programming with refinement operators~\cite{Lehmann2011Celoe}. These approaches start with the $\top$ concept and further refine it using a refinement operator \refine{} to generate new concepts. These newly created concepts are scored with respect to a scoring function, e.g., F1 measure, and the expression with the highest score that has not yet been refined before is then chosen to be refined further. This is repeated until either one of the found concepts achieves the maximum possible score or a given budget in the form of runtime or iterations has been consumed. Figure~\ref{fig:example-tree-1} shows an example of the search tree that is created. While this approach seems simple, it has been proven to be guaranteed to find a perfect solution for a given learning problem if 
\begin{enumerate*}
    \item such a solution exists, and
    \item the refinement operator is weakly complete, i.e., is able to generate any concept starting from $\top$~\cite{Lehmann2007LearningALC}.
\end{enumerate*} 
However, the search may take a very long time since the search space itself is infinite, i.e., every concept created by a weakly complete refinement operator can be further refined to create new  expressions~\cite{Lehmann2007LearningALC}.
Hence, several optimizations have been proposed.
For example, the latest version of \celoe{} achieves a reduced runtime by storing the knowledge base $\mathcal{K}$ in a triple store and collecting the counts necessary for scoring using SPARQL~\cite{Bin2016}. The refinement operator of \dlfoil{}~\cite{Fanizzi2018DLFoil} does not generate all refinements but a random subset, reducing the number of generated concepts.
In a similar way, Rizzo~\etal~\cite{Rizzo2017Trees} use a refinement operator with random sampling to generate terminological decision trees and combine multiple of them, similar to a random forest.
Another optimization is to calculate an upper bound of the achievable performance of the refinements of a concept and discard concepts with an upper bound lower than the quality of the best solution found so far~\cite{Sherif2017Wombat}.
In contrast to the previous approaches, \drill{}~\cite{Demir2023Drill} does not rely on a pre-defined scoring function to choose the expression that should be further refined. Instead, it uses deep Q-learning to train an agent that makes this decision. Our approach \celsystem{} shares the usage of a downward length-based refinement operator and a pre-defined scoring function with some of these approaches. However, our approach prunes the search space by avoiding the generation of concepts that lead to a low performance.
This pruning allows us to achieve results even on larger knowledge bases, without any pre-training.

Not all concept learning approaches rely on a refinement operator.
\evolearner{}~\cite{Heindorf2022EvoLearner} uses biased random walks on the knowledge base to create a start population of concepts. After that, it uses an evolutionary algorithm to create new concepts from this population.
\nces{}~\cite{Kouagou2023NCES} tackles concept learning as a translation problem. It synthesizes a solution by using sets of positive and negative examples as input to a neural network. 


\section{Preliminaries}
\label{sec:preliminaries}

\subsection{Description Logics}
\label{sec:dl}
Description logics are a family of languages for knowledge representation~\cite{baader2003description}. Within this article, we focus on the description logic \alc{}. Table~\ref{tab:alc} below defines the syntax and semantics of the \alc{} constructs. 
\begin{table}[htb]
\caption{Syntax \& semantics for \alc{} concepts~\cite{Kouagou2023NCES}. $\interpretation$ stands for an interpretation with domain $\domain$.}
	\centering 
\resizebox{\columnwidth
 }{!}{
   \begin{tabular}{@{}lccc@{}}
        \toprule
		\textbf{Construct}               & \textbf{Syntax}         & \textbf{Semantics} \\
		\midrule
        Atomic concept          & \conceptC{}    & $\interpret{\conceptC}\subseteq{\domain}$ \\
        Atomic role             & $\role$        & $\interpret{\role}\subseteq{\domain\times \domain}$\\
  		Top concept             & $\top$         & $\domain$\\
  		Bottom concept          & $\bot$         & $\emptyset$            \\
 		
 		Negation                & $\neg \conceptC$& $\domain\setminus \interpret{\conceptC}$
 		\\
 		Conjunction             & $\conceptC \sqcap \conceptD$    & $\interpret{\conceptC} \cap \interpret{\conceptD}$ \\
 		Disjunction             & $\conceptC \sqcup \conceptD$    & $\interpret{\conceptC} \cup \interpret{\conceptD}$\\
 		Existential restriction & $\exists \role.\conceptC$ & $\{ x \mid \exists b: (a,b) \in 
   \interpret{\role} \land b \in \interpret{\conceptC}\}$\\
 		Universal restriction & $\forall \role.\conceptC$   & $\{ a \mid \forall b: (a,b) \in \interpret{\role} \implies b \in \interpret{\conceptC}\} $\\
		\bottomrule
	\end{tabular}
}

 \label{tab:alc}
\end{table}

\subsection{Refinement Operators}
\label{sec:refinement-operators}

A quasi-ordering is a reflexive and transitive relation~\cite{Lehmann2007Refinement}. 
Let \mbox{$(\concepts{},\preccurlyeq)$} be a quasi-ordered space. A downward refinement operator $\refine$ in such a space is a mapping from $\concepts$ to $2^{\concepts}$ such that $\forall \conceptC \in \concepts : \conceptD \in \refine(\conceptC) \implies \conceptD \preccurlyeq \conceptC$. \conceptD{} is called a specialisation of \conceptC{}~\cite{Lehmann2007Refinement}.

Two quasi-orderings are often used in concept learning. Some of the earliest approaches rely on the subsumption relation $\sqsubseteq$~\cite{Lehmann2011Celoe}. More recent approaches (including ours) use the length of concepts, which is defined recursively for \alc{} concepts as follows~\cite{Kouagou2022CLIP}:
\begin{equation}
\length(\conceptC)=\begin{cases}
1 & \text{if } \conceptC\!\in\! \namedConcepts{} \cup \{\top, \bot \},\\
1 + \length(\conceptX) & \text{if } \conceptC\!=\!\neg \conceptX, \\
2 + \length(\conceptX) & \text{if } \conceptC\!\in\!\{\exists r.\conceptX,\forall r.\conceptX\},\\
1 + \length(\conceptX) + \length(\conceptY) & \text{if } \conceptC\!\in\!\{\conceptX \sqcap \conceptY, \conceptX \sqcup \conceptY\}.\\
\end{cases}
\end{equation}

\subsection{Concept Learning}
\label{sec:concept-learning}
Let \namedIndividuals{} (individuals), \namedRoles{} (roles), and \namedConcepts{} (named concepts) be infinite, countable and pairwise disjoint sets. A knowledge base $\knowledgebase=(\tbox,\abox)$ over a description logic $\mathcal L$ is a pair that consists of a T-Box $\tbox$ and an A-Box $\abox$. The T-Box contains subsumption axioms of the form $\conceptC \sqsubseteq \conceptD$, where \conceptC{} and \conceptD{} are concepts in $\mathcal L$, e.g., \alc{} (see Section \ref{sec:dl}). The A-Box contains assertions of the form $\conceptC(a)$ or $\role(a,b)$, where $\conceptC$ is a concept in $\mathcal L$, $\role \in \namedRoles{}$ is a role, and $a,b \in \namedIndividuals{}$ are individuals~\cite{Kouagou2023NCES}.

A concept learning problem over a knowledge base \knowledgebase{} consists of a pair $\examples{}=(\pexamples,\nexamples)$, where $\pexamples \subseteq \namedIndividuals$ is the set of positive examples and $\nexamples \subseteq \namedIndividuals$ contains negative examples. The goal of concept learning is to find a concept $\conceptC$ which satisfies~\cite{Demir2023Drill}:

\begin{equation}
\label{eq:conceptlearning}
\forall \Example \in \pexamples: \knowledgebase \models \conceptC(\Example) \land \forall \Example \in \nexamples: \knowledgebase \not\models \conceptC(\Example).
\end{equation}

Finding such a concept is not always possible. Hence, most concept learners aim to maximize a quality function $\quality: \concepts \rightarrow [0,1]$, where $\concepts$ is the set of all concepts over $\namedConcepts{}$ and $\namedRoles{}$
in $\mathcal L$. Quality functions, e.g., F1 measure, are commonly designed to return 1 for inputs that satisfy Equation~\ref{eq:conceptlearning} (see~\cite{Kouagou2023NCES2}, Definition 3).

\subsection{Knowledge-Graph-based Question Answering}
A knowledge graph \referenceGraph{} is ``a graph of data intended to accumulate and convey knowledge of the real world, whose nodes represent entities of interest and whose edges represent relations between these entities''~\cite{hogan2021knowledge}.
Given a knowledge graph \referenceGraph{} and an input question \Question{} in natural language, the goal of a knowledge-graph-based question answering (QA) system \qasystem{} is to derive the set of answers \answers{} for \Question{} using \referenceGraph{}~\cite{ji-etal-2024-retrieval}.
We formalize this as:
\begin{align}
    \label{eq:kgqa_sparql_generation}
    \answers &= \qasystem(\referenceGraph, \Question)\,.
\end{align}

\section{Explainable Benchmarking}

Our goal is to automatically generate explanations for benchmarking results that provide users and developers with insights that allow them to better use, or improve the benchmarked system \cite{rohlfing2020explanation}. As a running example, let's assume there is a QA system that answers questions related to geography well but others subpar. The goal of our work is to give this insight into the system's performance in an automatic way based on evaluation results.
Our approach goes beyond previous analysis tools that have been designed for a particular field as it does not rely on pre-defined features that are bound to a specific use case. 
Instead, we transform the available information about the benchmarking process into structured data and apply concept learning.
Hence, our approach only has the requirement that the available data can be transformed into structured data and that the used concept learning algorithm is expressive enough to find an explanation.

\subsection{Approach}

\begin{figure}
    \centering

    \includegraphics[width=\linewidth]{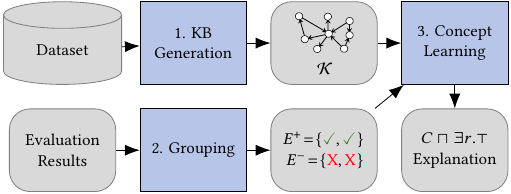}
    
    \caption{Overview over the three steps of our approach.}
    \Description{The figure shows a data flow diagram. It starts with a dataset from which an arrow goes into "step 1 knowledge base generation". The output of this step is a knowledge base K. In parallel, the evaluation results go into "step 2 grouping". The result of this step are the two sets E+ and E-. These sets and the knowledge base K are the input to step "3 concept learning". The result of this step is a concept which represents the explanation.}
    \label{fig:overview}
\end{figure}

Our approach to automatically generate explanations for benchmarking results consists of the three steps shown in Figure~\ref{fig:overview}: \begin{enumerate*}
    \item generate a knowledge base (\knowledgebase{}) comprising structured information about the content of the benchmark dataset,
    \item split the benchmark's tasks, e.g., the questions of a QA dataset, into correctly (\pexamples{}) and incorrectly answered tasks (\nexamples{}), and
    \item use concept learning to determine an expression that separates the two groups from each other.
\end{enumerate*}
We will explain these steps in more detail in the following.

\label{sec:approach}

\begin{figure}
     \centering
     \includegraphics[width=\linewidth]{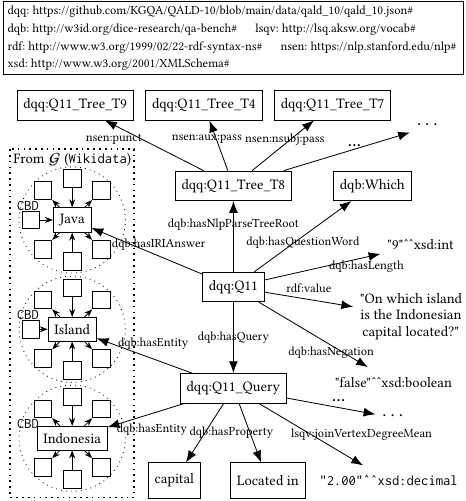}
     \caption{An excerpt of the knowledge base around the question \texttt{dqq:Q11} (question 11 from QALD-10). Wikidata entities and properties are replaced by their labels.}
     \Description{The figure shows a small excerpt of a knowledge base. In the center, is a node representing the question Q1. Its natural language form starts with "On which island is the Indonesian capital located?" and is connected to the question with rdf:value. The question has the answer "Java", which is an entity from the reference graph G on which the questions should be answered. Together with the entity Java, its CBD is part of this knowledge base. Another entity connected to the question is the SPARQL query, that if executed on the reference knowledge graph G would provide the correct answer set. The query is connected to the properties "Located in" and "Capital" that occurs in the query with the dqb:hasProperty property. The query also has a triple expressing that its lsqv:joinVertxDegreeMean is 2.00. The query also contains the entity "Java", which is present in the knowledge base including its CBD gathered from the reference knowledge graph G and it is connected to the query with the dqb:hasEntity property. In the upper part of the figure, there are several entites that are connected with properties from a stanford NLP namespace. They represent a dependency parse tree generated using the Stanford NLP library. The root of the tree is connected to the question entity with the dqb:hasNlpParseTreeRoot property.}
     \label{fig:example-question-graph}
\end{figure}

In the first step, we generate a knowledge base comprising information from the dataset about the examples that are used during the benchmarking process (e.g., questions in the area of QA). The more information can be provided about the examples, the higher is the chance that the concept learning algorithm is going to find a good explanation in step 3. For the application of our approach on QA benchmarking results, we rely on the QALD datasets~\cite{perevalov2022qald,qald10}, from which we extract features based on:
\begin{enumerate*}
    \item the natural language question (e.g., length, question word, dependency parse tree~\cite{manning-etal-2014-stanford}), 
    \item the ground truth answer(s) (e.g., type of answers, concise bounded description (CBD)~\cite{cbd_w3c_2005} of entities in the answer set gathered from \referenceGraph{}), and
    \item the provided SPARQL query, that returns the ground truth answer when used on \referenceGraph{} (e.g., LSQ~\cite{Saleem2015} features, properties and CBDs of entities used the query).\footnote{Our Github project provides a more detailed description of the information.} 
\end{enumerate*}
Figure~\ref{fig:example-question-graph} shows the example question \texttt{dqq:Q11} (``On which island is the Indonesian capital located?'') and an excerpt of the data in \knowledgebase{} connected to it.


\begin{figure}[tb]
    \centering
     \includegraphics[width=\linewidth]{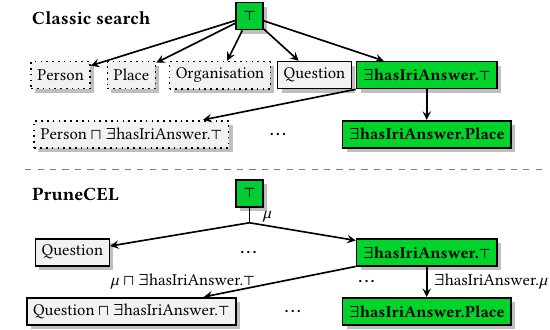}\\
    
    \caption{Examples of a classic (top) and a \celsystem{} (bottom) search tree. The green expressions are further refined. Expressions with a dotted frame do not have any given example as instance. The labels on the edges are the generated templates used to derive concepts or roles from the oracle.}
    \Description{The figure shows two examples of search trees. The upper tree is generated by a typical top-down refinement operator. The tree starts with the most generic expression Top. This is refined to several concepts. Some of them like Person, Place and Organization are marked with a dotted frame, expressing that these concepts don't have any of the given positive or negative examples as instances. The concept Question and exists hasIriAnswer Top are also created. The latter is marked green, meaning that it is chosen for further refinement. This refinement leads to a new set of concepts from which only two are shown. One is again marked green but the other one is marked with a dotted frame.
    The lower tree in the figure shows a search tree generated by PruneCEL. It is similar to the previous one. However, this time, patterns are written on the edges and there are no concepts with dotted frames, i.e., all the concepts have at least one of the given positive or negative examples as instance. The patterns can be best understood with the concept refinement of exists hasIriAnswer Top. One arrow shows a conjunction of µ and exists hasIriAnswer Top. This means that the search will look for something that fits into this conjunction, i.e., into the position of µ and creates a concept that selects at least one of the examples. Another arrow has the pattern exists hasIriAnswer µ, leading to concepts that has a named concept or a role instead of top in the position of µ. This arrow leads to the example result exists hasIriAnswer Place.}
    \label{fig:example-tree-1}
\end{figure}

In the second step, we separate the questions into two groups depending on the ability of the question answering system to answer the questions correctly.
We determine the correctness of each question using GERBIL QA~\cite{usbeck2019benchmarking} and define that a question is answered correctly if its F1 score is 0.5 or higher.
Conversely, questions with an F1 score below 0.5 are labeled as incorrectly answered.
We use the correctly answered questions as $E^{+}$ and incorrectly answered questions as $E^{-}$ as defined in Section~\ref{sec:preliminaries}. In our running example, we assume that the evaluated QA system is able to answer the question from Figure~\ref{fig:example-question-graph} and other geography-related questions correctly. Hence, they are part of $E^{+}$ while other questions that were not answered well are assigned to $E^{-}$.

In the third step, we learn a concept that characterizes the previously identified positive examples $E^{+}$ and separates them from the negative examples $E^{-}$.
This concept is used as explanation and can be easily verbalized to be human-readable~\cite{Vollmers2024Verbalize}.
In our running example, a learner could return the concept $\exists\text{hasIriAnswer.}\text{Place}$ which can be verbalized to ``The system can answer questions that have places as answers'' providing insight into the QA system's performance in a human-readable form.
In this step, any concept learning approach can be used. 
However, we propose a new, scalable approach in the following.

\subsection{PruneCEL}

\celsystem{} is a top-down refinement-operator-based concept learning algorithm. Unlike similar approaches like \textsc{CELOE} or \textsc{DRILL} (see Section~\ref{sec:related-work}), \celsystem{} avoids the generation of unsatisfiable concepts. Correspondingly, it is more time-efficient than similar approaches. The motivation behind our approach lies in using the monotonicity of subset inclusion, i.e., $A \subset B \implies |A| < |B|$. Hence, if $\refine$ is a downward refinement operator, then $\forall \conceptD \in \refine(\conceptC): \interpret{\conceptC} \cap (\pexamples \cup \nexamples) = \emptyset \implies \interpret{\conceptD} \cap (\pexamples \cup \nexamples) = \emptyset$.
While other approaches exploit this observation by aiming not to refine concepts $\conceptC$ with $\interpret{\conceptC} \cap (\pexamples \cup \nexamples) = \emptyset$, the main idea of \celsystem{} is to avoid generating such concepts entirely.


The upper half of Figure~\ref{fig:example-tree-1} shows a typical search tree as it would be created for our running example. The top concept is refined to a long list of concepts. The concept with the best performance $\exists\text{hasIriAnswer.}\top$ is then further refined to another long list of concepts.
In practice, many of the generated concepts do not include any of the given positive or negative examples as instances. For the running example, the system may generate a concept such as $\exists\text{hasIriAnswer.Person}$, which does not describe the positive geography-related questions but still consumes resources during its generation and evaluation.

\celsystem{} avoids the generation of these concepts by relying on an oracle, which provides named concepts or roles that can be used to fill certain gaps in a template, generating concepts that have at least one of the given examples as instance.


When refining the concept $\exists\text{hasIriAnswer.}\top$ in our running example, \celsystem{} generates templates like $ \exists\text{hasIriAnswer.}\marker$ including the concept to be refined and a marker $\marker$. Then, it uses an oracle to get all named concepts or roles that can be used to replace $\marker$ to generate new concepts. The oracle guarantees that for each generated concept $\conceptD$ the following holds: $\interpret{\conceptD} \cap (\pexamples \cup \nexamples) \neq \emptyset$. In our example, the oracle suggests to replace $\marker$ with \text{Place} in to produce a meaningful candidate concept.
The lower half of Figure~\ref{fig:example-tree-1} shows the search tree for the same example as \celsystem{} would create it.

This take on concept learning leads to several changes in the typical recursive workflow of the refinement operator.
First, our operator has to be able to generate templates. Each of them contains exactly one marked position at which the oracle should insert named concepts or roles to create a new concept.
Second, we use SPARQL queries to implement the oracle, i.e., to select named concepts and roles from \knowledgebase{}. In the following, we provide a formal definition of our refinement operator before we provide details about \celsystem{}'s oracle, scoring and extensions.

\subsubsection{Refinement Operator}
We define a top-down length-based refinement operator $\refine: \concepts \times \exampleSpace \times \exampleSpace \rightarrow 2^{\concepts}$ for \alc{}. \refine{} operates in the quasi-ordered space $(\concepts, \length)$, where $\length : \conceptC \rightarrow \mathbb{N}$ is the length of a concept as defined above. 
Hence, $\conceptD \in \rho(\conceptC) \implies \length(\conceptD) \geq \length(\conceptC)$.

Let \templates{} be the space of all templates that can be created when extending the \alc{} grammar defined in Section~\ref{sec:dl} with the symbol $\marker$. $\marker$ is handled like a named concept but serves as a marker of the position within a template that has to be filled by the oracle. We define that each template in \templates{} contains $\marker$ exactly once. Further, let $\tmerge: \templates \times \templates \rightarrow \templates$ be a function that merges two given templates $\template_1$ and $\template_2$ by replacing the occurrence of $\marker$ in the template $\template_1$ with the template $\template_2$. Similarly, let $\cmerge: \templates \times \concepts \rightarrow \concepts$ be a function that creates a new concept by replacing the marker $\marker$ in the given template with the given concept.

Based on the previous definitions, we define our top-down length-based refinement operator $\refine$ to refine a given concept $\conceptC$ based on the given examples $\pexamples$ and $\nexamples$ as follows:

\begin{equation}
\label{eq:rho}
\begin{split}
\rho(\conceptC, \pexamples, \nexamples) =& \rho^{\star}(\conceptC, \marker, \pexamples, \nexamples) \cup\\
&\!\left\{(\neg \conceptD \left| \conceptD\in\rho^{\star}(\conceptC, \marker, \pexamples, \nexamples)\right.\right\},
\end{split}
\end{equation}

where $\trefine: \concepts \times \templates \times \exampleSpace \times \exampleSpace \rightarrow 2^{\concepts}$ is a function that takes a concept and a template and refines it recursively based on the given positive and negative examples. Since Equation~\ref{eq:rho} is the start of the recursion, the template only comprises the positional marker $\marker$. It can also be seen that the refinement operator doubles the amount of expressions that \trefine{} provides by creating their negations. 
We define the recursive function \trefine{} as follows:
\begin{equation}
\label{eq:rho-star}
\begin{split}
&\rho^{\star}(\conceptC, \template, \pexamples, \nexamples)\\
&=\!\begin{cases}
\rho^{\star}(\conceptX,\tmerge(\template,\exists r.\marker), \pexamples, \nexamples)\cup
&\!\!\text{if\,}\conceptC\!\!=\!\!\exists r.\conceptX,\!\\
\quad \{ \cmerge(\template,\forall r.\conceptX) \} \\
\rho^{\star}(\conceptX,\tmerge(\template,\forall r.\marker), \pexamples, \nexamples)
&\!\!\text{if\,}\conceptC\!\!=\!\!\forall r.\conceptX,\!\\
\rho^{\star}(\conceptX, \tmerge(\template,\neg \marker), \pexamples, \nexamples)
&\!\!\text{if\,}\conceptC\!\!=\!\!\neg \conceptX \land \conceptX\!\notin \!\namedConcepts,\!\\
\rho^{\star}(\conceptX,\tmerge(\template, \marker\!\sqcap\!\conceptY), \pexamples, \nexamples) \cup 
&\!\!\text{if\,}\conceptC\!\!=\!\!\conceptX\!\sqcap\!\conceptY,\!\\
\quad \rho^{\star}(\conceptY, \tmerge(\template, \conceptX\!\sqcap\!\marker), \pexamples, \nexamples) \\
\rho^{\star}(\conceptX, \tmerge(\template, \marker\!\sqcup\!\conceptY), \pexamples, \nexamples) \cup
&\!\!\text{if\,}\conceptC\!\!=\!\!\conceptX\!\sqcup\!\conceptY,\!\\
\quad \rho^{\star}(\conceptY, \tmerge(\template, \conceptX\!\sqcup\!\marker), \pexamples, \nexamples) \\
\generator(\tmerge(\template, \conceptC\!\sqcap\!\marker), \pexamples, \nexamples)\cup
&\!\!\text{if\,}\conceptC\!\!\notin\!\!\{\!\top,\bot\!\},\!\\
\quad \generator(\tmerge(\template, \conceptC\!\sqcup\!\marker), \pexamples, \nexamples)\\
\generator(\template, \pexamples, \nexamples)
&\!\!\text{if\,}\conceptC\!\!=\!\!\top\,,
\end{cases}
\end{split}
\end{equation}
where $\conceptX,\conceptY\in \concepts$ and $\generator$ is a function that generates new concepts using the oracle and the given template.
For complex concepts (the first five cases in Equation~\ref{eq:rho-star}), \trefine{} calls itself recursively, focusing on one of the parts of the given concept while the other parts of the expression are added to the template using the merge function \tmerge{}.
An exception is the first case, in which $\exists r.\conceptX$ is also refined to $\forall r.\conceptX$. 
For all expressions except $\top$ or $\bot$ (case 6), our refinement tries to add a conjunction and a disjunction. The $\top$ expression is replaced with named classes or roles (case 7). The two latter cases are the base cases of the recursion and rely on the generator function $\generator$.

The generator function $\generator: \templates \times \exampleSpace \times \exampleSpace \rightarrow 2^{\concepts}$ takes a template and the examples as input and returns new concepts based on the named concepts and roles derived from the oracle.
We define it as follows:
\begin{equation}
\label{eq:generator}
\begin{split}
\!\!\!\generator(\template, \pexamples, \nexamples)=
&\{\cmerge(\template, \conceptD) | \conceptD \in \selconcept(\template, \pexamples, \nexamples)\} \cup\\
&\{\cmerge(\template,\neg \conceptD) | \conceptD \in \selnegconcept(\template, \pexamples, \nexamples)\} \cup\!\\
&\{\cmerge(\template,\exists \role.\top) | \role \in \selrole(\template, \pexamples, \nexamples)\}\,,
\end{split}
\end{equation}
where we rely on three different functions---\selconcept{}, \selnegconcept{}, and \selrole{}---provided by the oracle. The first oracle function $\selconcept: \templates \times \exampleSpace \times \exampleSpace \rightarrow \namedConcepts$ takes a template and a set of positive and negative examples as input. 
The output is a set comprising all named concepts that when used to replace the marker \marker{} in the template form a concept \conceptD{} that has at least one of the given examples as instance:

\begin{equation}
\begin{split}
\selconcept(\template, \pexamples, \nexamples) &= \{\conceptD | \conceptX=\cmerge(\template,\conceptD) \land \conceptD \in \namedConcepts \\
&\land \interpret{\conceptX} \cap (\pexamples \cup \nexamples) \neq \emptyset\}\,.
\end{split}
\end{equation}

In the same way, we define the second function $\selnegconcept: \templates \times \exampleSpace \times \exampleSpace \rightarrow \namedConcepts$, which negates the named concepts as follows:
\begin{equation}
\begin{split}
\selnegconcept(\template, \pexamples, \nexamples) &= \{\conceptD | \conceptX=\cmerge(\template,\neg \conceptD) \land \conceptD \in \namedConcepts \\
&\land \interpret{\conceptX} \cap (\pexamples \cup \nexamples) \neq \emptyset\}\,.
\end{split}
\end{equation}

Similarly, we define the third oracle function $\selrole: \templates \times \exampleSpace \times \exampleSpace \rightarrow \namedRoles$ that returns roles as follows:
\begin{equation}
\begin{split}
    \selrole(\template, \pexamples, \nexamples) &= \{ \role | 
    \conceptX=\cmerge(\template, \exists \role.\top) \land \role \in \namedRoles \\
&\land \interpret{\conceptX} \cap (\pexamples \cup \nexamples) \neq \emptyset\}\,.
\end{split}
\end{equation}

According to~\cite{Lehmann2007Refinement}, $\rho$ is a top-down length-based refinement operator since Equation~\ref{eq:rho-star} guarantees that newly created concepts have either the same length (cases 1 and 7) or are longer (case 6) than the given concept $C$, i.e., $\conceptD \in \refine(\conceptC) \implies \length(\conceptD) \geq \length(\conceptC)$.
From the list of properties of a refinement operator proposed by ~\cite{Lehmann2007Refinement}, our refinement operator $\refine$ is finite (i.e., $\refine(\conceptC)$ is finite for any concept $\conceptC$) and redundant (i.e., during the search, \refine{} may return a concept that is equivalent to a previously returned concept). However, \refine{} is not (weakly) complete as it is not able to generate all possible concepts since many do not select any given example.
Hence, pruning the search space leads to the loss of the completeness of the operator.\footnote{A similar tradeoff is encountered by other approaches, e.g., \dlfoil{}~\cite{Fanizzi2018DLFoil}. The interested reader is referred to \cite{Lehmann2007Refinement} for the full list of properties.}

\subsubsection{Oracle Implementation}

The oracle is implemented in the form of SPARQL queries that are sent to a triple store containing \knowledgebase{}. Since the queries used for \selconcept{}, \selnegconcept{}, and \selrole{} already contain the positive and negative examples, we extend these queries to derive the numbers of positive and negative examples that are instances of the newly created concepts.
This further reduces the number of SPARQL queries that our approach sends to the triple store in comparison to previous approaches, which would derive these counts for all created concepts one after the other.

\subsubsection{Heuristic}
We score a generated class expression $\conceptC$ by determining the number of positive \poscount{} (respectively, negative \negcount{}) examples that are instances of (respectively, ruled out by) this concept according to \knowledgebase{}. Then, we compare these counts with the overall number of positive and negative examples. 
This can be done using any quality function \quality{} like accuracy or F1 measure. Like previous works, e.g.,~\cite{Heindorf2022EvoLearner}, we assume that shorter concepts are more general and, hence, preferred. So we include the length of the concept multiplied by a small constant $\eta$ into our heuristic function \measure{}:
\begin{equation}
\measure(\conceptC, \poscount, \negcount, \pexamples, \nexamples) = \quality(\poscount, \negcount,|\pexamples|,|\nexamples|) - \eta \length(\conceptC)\,.
\end{equation}

\subsubsection{Extensions}

We propose two additional extensions---\celsystem{}-S and \celsystem{}-R. Both can be used together, which we name \celsystem{}-RS.

\paragraph{\celsystem{}-S} In this mode, a newly created concept $\conceptD \in \refine(\conceptC)$ is only considered for further refinement if \begin{enumerate*}
    \item it received a better score than $\conceptC$ or
    \item $\conceptD$ has been derived from $\conceptC$ by adding a role.
\end{enumerate*}

\paragraph{\celsystem{}-R} In this recursive mode, \celsystem{} calls itself if it found a solution for a sub-problem. Let $\subpexamples \subset \pexamples$ be a set of positive examples with $|\subpexamples| \geq 2$. If our approach has found a concept $\conceptD$, which is an exact solution for $\subpexamples$, i.e.,  $\conceptD$ satisfies $\forall \Example \in \subpexamples: \knowledgebase \vDash \conceptD(\Example) \land \forall \Example \in \nexamples: \knowledgebase \nvDash \conceptD(\Example)$, \celsystem{} calls itself with a smaller learning problem $(\pexamples\backslash \subpexamples, \nexamples)$. It spends a limited amount of iterations 
on this smaller problem before it returns its best concepts. These are combined with $\conceptD$ and introduced into the search tree as additional solutions which then can be further refined.

\section{Evaluation}

\subsection{Experiment Setup}
First, we compare the performance of different concept learners on the knowledge bases created by our approach (Experiment I). 
Finally, we use a survey to check whether our explanations are understood by humans (Experiment II).




\begin{table}[tb]
\centering
\caption{The number of learning problems (LP), their average number of positive and negative examples, and the features of \knowledgebase{} for the QALD-based datasets. P = Properties.}
\begin{tabular}{@{}l@{}r@{\ }r@{\ }r@{\ \ }r@{\ \ }r@{\ \ }r@{}}
\toprule
\multicolumn{1}{@{}c@{}}{\textbf{Datasets}} &
\multicolumn{1}{@{}c@{\ }}{\textbf{LPs}} & 
\multicolumn{1}{@{}c@{\ }}{\(\boldsymbol{|E^{+}|}\)}& 
\multicolumn{1}{@{}c@{}}{\(\boldsymbol{|E^{-}|}\)}& 
\multicolumn{1}{@{}c@{}}{\textbf{Entities}} & 
\multicolumn{1}{@{}c@{}}{\textbf{P}} & 
\multicolumn{1}{@{}c@{}}{\textbf{Triples}} \\
\midrule
QALD9+DB        & 3  & 22.7 & 110.3 & 21,518,759 & 918 & 72,737,644 \\
QALD9+WD        & 2  & 30.5 & 85.5 & 36,565,453 & 826 & 84,345,960 \\
QALD10          & 2  & 91.0 & 303.0 & 64,352,096 & 878 & 155,959,524 \\
\bottomrule
\end{tabular}

\label{tab:dataset-stats}
\vspace{-2mm}
\end{table}


\subsubsection{Experiment I}

\begin{table}[tb]

\centering
\caption{Correctly / faulty answered questions per QA system. DB = DBpedia, WD = Wikidata.}
\setlength{\tabcolsep}{3pt} 
\begin{tabular}{@{}lccc}
\toprule
\multicolumn{1}{c}{\textbf{Systems}} &
\textbf{QALD9+DB} &
\textbf{QALD9+WD} &
\textbf{QALD10} \\
\midrule
\deeppavlov{}& -- / --\phantom{1} & 26 / 90 & \phantom{1}61 / 333 \\
\ganswer{}   & 18 / 115 & -- / -- &  -- /  -- \\
\mst{}       & 28 / 105 & 35 / 81 & 121 / 273 \\
\tebaqa{}    & 22 / 111 & -- / -- &  -- /  -- \\
\bottomrule
\end{tabular}

\label{tab:qa-stats}
\vspace{-2mm}
\end{table}

In the First experiment, we apply our approach to the benchmarking results of the \qasystems{} QA systems \deeppavlov{}~\cite{Zharikova2023deeppavlov}, \ganswer{}~\cite{Hu2018ganswer}, \tebaqa{} \cite{Vollmers2021tebaqa}, and \mst{}~\cite{srivastava2024mst5} on three QA datasets---QALD 9 Plus for DBpedia and Wikidata~\cite{perevalov2022qald}, and QALD 10~\cite{qald10}.
We remove questions that have an empty ground truth answer set from these QA datasets leading to 133, 116, and 394 questions, respectively.
We generate a knowledge base \knowledgebase{} for each QA dataset as described in Section~\ref{sec:approach}.
We gather the answers generated by the \qasystems{} QA systems for these datasets.
The DBpedia-based systems \ganswer{} and \tebaqa{} only provide answers for QALD 9 Plus DBpedia, while the Wikidata-based system \deeppavlov{} provides results for QALD 9 Plus Wikidata and QALD 10.
\mst{}~\cite{srivastava2024mst5} provides answers for all three QA datasets.
We use the evaluation results from GERBIL QA~\cite{usbeck2019benchmarking} to identify correctly and faulty answered questions to derive \pexamples{} and \nexamples{} for each QA system.
Table~\ref{tab:qa-stats} shows the summary of this step. Table~\ref{tab:dataset-stats} summarizes the features of the generated knowledge bases as well as the resulting concept learning datasets dubbed QALD9+DB, QALD9+WD, and QALD10.\footnote{The DBpedia and Wikidata versions that we use as \referenceGraph{} can be found at  \url{https://downloads.dbpedia.org/2016-10/core-i18n/en/} and~\cite{Qald10kg2022}.} 

On these three concept learning dataset, we apply the four concept learners \celoe{}, \drill{}, \evolearner{}, and \nces{} from the related work. We compare their performance to \celsystem{}-RS, which showed the best performance in preliminary experiments.\footnote{Due to the length restriction of this publication, the results of these preliminary experiments can be found in our Github project.} 
We run all approaches with their default configuration.\footnote{For \drill{}, we use the Keci embedding algorithm~\cite{demir2023clifford}. The embedding models and pre-trained \drill{} models can be found at \doi{10.5281/zenodo.14720609} and \doi{10.5281/zenodo.14720524}, respectively.}
\celsystem{}-RS is executed three times using three different measures \measure{} as part of the scoring function, namely accuracy, balanced accuracy and F1 measure. In all configurations, we set $\eta=0.01$.
We set the maximum runtime of all approaches for a single learning problem to 10 minutes and compare their results using the F1-measure, the concept length and their runtime.\footnote{We provide the knowledge bases and learning problems at \doi{10.5281/zenodo.14720669} and \doi{10.5281/zenodo.16681824}, respectively. We use an AMD EPYC 7282 with 252 GB RAM.}

\subsubsection{Experiment II}

\begin{table*}[tb]
\centering
\caption{F1 score, length of the generated concepts ($\length(\conceptC)$), and runtime (RT, in seconds) for the learning problems (LP) of Experiment II. For \celsystem{}-RS, we report the quality measure ($\quality{}$, A = accuracy, B = balanced accuracy, F = F1 measure) leading to the best F1 score. The complete set of results for \celsystem{}-RS can be found in the appendix.}
\resizebox{\linewidth}{!}{
\begin{tabular}{@{}l@{\ }l@{\ }r@{}c@{}r@{\ }c@{}c@{}cr@{}c@{}rc@{}c@{}cr@{}c@{}r@{}c@{}r@{}}
\toprule
\multicolumn{2}{c}{\textbf{Dataset}} 
& \multicolumn{3}{c}{\textbf{Drill}} 
& \multicolumn{3}{c}{\textbf{Evolearner}} 
& \multicolumn{3}{c}{\textbf{CELOE}} 
& \multicolumn{3}{c}{\textbf{NCES}} 
& \multicolumn{4}{c}{\textbf{\celsystem{}-RS}} \\
\cmidrule(r){1-2}
\cmidrule(lr){3-5}
\cmidrule(lr){6-8}
\cmidrule(lr){9-11}
\cmidrule(lr){12-14}
\cmidrule(l){15-18}
\multicolumn{1}{c}{\knowledgebase{}} & 
\multicolumn{1}{c}{LP} & 
\multicolumn{1}{c}{F1} & 
\multicolumn{1}{c}{$\length(\conceptC)$} & 
\multicolumn{1}{c}{RT} & 
\multicolumn{1}{c}{F1} & 
\multicolumn{1}{c}{$\length(\conceptC)$} & 
\multicolumn{1}{c}{RT} & 
\multicolumn{1}{c}{F1} & 
\multicolumn{1}{c}{$\length(\conceptC)$} & 
\multicolumn{1}{c}{RT} & 
\multicolumn{1}{c}{F1} & 
\multicolumn{1}{c}{$\length(\conceptC)$} & 
\multicolumn{1}{c}{RT} & 
\multicolumn{1}{c}{F1} & 
\multicolumn{1}{c}{$\length(\conceptC)$} &
\multicolumn{1}{c}{RT} &
\multicolumn{1}{c}{\quality{}}  \\
\midrule

QALD9+DB & \ganswer{} & 0.24 & 1 & 2299.2  
& -- & -- & --
& \underline{0.26} & 3 & 10306.2 
& -- & -- & -- 
& \textbf{0.35} & \phantom{1}19 & 600.1 & A \\ 

& \mst{} & \underline{0.35} & 1  & 2694.5
& -- & -- &  --
& \underline{0.35} & 1 & 10392.3 
& -- & -- & -- 
& \textbf{0.57} & \phantom{1}24 & 600.1 & B \\ 

& \tebaqa{} & \underline{0.30} & 3 & 2267.4  
& -- & -- & -- 
& \underline{0.30} & 3 & 10356.7 
& -- & -- & -- 
& \textbf{0.44} & \phantom{1}23 & 650.5 & F \\ 

QALD9+WD 
& \deeppavlov{} & 0.37 & 1  & 3793.6
& -- & -- & -- 
& \underline{0.41} & 3 & 12416.7
& -- & -- & -- 
& \textbf{0.96} & 107 & 600.9 & A \\ 

& \mst{}& 0.46  & 1 & 3779.8
& -- & -- & --
& \underline{0.50} & 3 & 12369.0
& -- & -- & --
& \textbf{0.84} & \phantom{1}28 & 600.9 & B \\ 

QALD10 
& \deeppavlov{} & \underline{0.27} & 1  & 3495.3  
& -- & -- & -- 
& \underline{0.27} & 1 & 2326.2 
& -- & -- & -- 
& \textbf{0.34} & \phantom{1}10 & 600.8 & F \\ 

& \mst{} & \underline{0.47} & 1 & 3468.1
& -- & -- & --
& \underline{0.47} & 1 & 2271.3
& -- & -- & --
& \textbf{0.56} & \phantom{1}20 & 602.0 & B \\ 

\bottomrule
\end{tabular}
}
\label{tab:results-exp2}
\vspace{-3mm}
\end{table*} 

We conduct a survey to evaluate the quality of our explanations.
We choose two concepts learned from Experiment I generated on two very different knowledge bases---QALD9+DB and QALD10---that achieve the highest F1 score when compared with a baseline that returns the concept $\top$. On both knowledge bases, these are concepts explaining the performance of \mst{}, which we verbalize using ChatGPT.\footnote{The prompt for the verbalization can be found in our Github project.}
Our approach explains the performance of \mst{} on QALD10 with the following concept:
\begin{align*}
  &\exists \text{hasEntityAnswer}.(\text{album} \sqcup 
  \exists \text{copyrightStatusAsCreator}.\top \sqcup \\ &\text{profession} 
\sqcup \exists \text{locatedInAdministrativeTerritorialEntity}.
  \neg \text{country} \sqcup \\
 & \exists \text{manifestationOf}.\top) 
  \sqcup \exists \text{hasBooleanAnswer}.\top
\end{align*}
which is verbalized as (naming \mst{} "QAS1"):\\
\textit{The system "QAS1" can answer questions if:}
\begin{enumerate}
    \item \textit{There’s an answer involving an album, a creator's copyright status, a profession, a location that’s not a country, or something that has a type or form.}
    \item \textit{Or, it can answer questions that have a simple yes/no (boolean) answer.}
\end{enumerate}

For \mst{} on QALD9+DB, our approach finds the following concept:
\begin{align*}
  &\exists \text{hasEntityAnswer}.((\neg\text{agent}\!\sqcap\!\exists\text{parentMountainPeak}.\!\top)\!\sqcup\! \text{building})\\
  & \sqcup (\exists \text{hasIRIAnswer}.(\text{astronaut} \sqcup (\neg\text{agent}\sqcap\neg\text{spatialThing}))\sqcap\\
  & \quad \exists\text{hasQuestionWord}.\!\top)
\end{align*}
which is verbalized as (naming \mst{} "QAS2"):\\
\textit{The system "QAS2" can answer questions if:}
\begin{enumerate}
    \item \textit{The answer involves either: a non-agent (not a person or entity with intent) with a parent mountain peak, or a building.}
    \item \textit{Or, if the answer involves: an astronaut, or a non-agent, non-spatial entity (something that’s neither a person nor a physical location), and if the question includes a question word (like "who," "what," or "where").}
\end{enumerate}
For each chosen concept, we randomly choose 5 correctly and 5 faulty answered questions of \mst{}, which are classified correctly by the chosen concept.
In the survey, we provide these 20 questions together with 5 or more statements from \knowledgebase{} that would be sufficient for a reasoner to decide whether the question with this data is an instance of the learned concept.
For each question, the survey participants have to decide whether a QA system with the provided explanation would be able to answer the given question.
The higher the success rate, the better do humans understand the explanation and are able to decide whether to use the QA system for it or not.
We configure the survey to randomly order the questions for each participant and distribute the survey to computer scientists (e.g., via mailing lists).

\subsection{Results}




\subsubsection{Experiment I}

Table~\ref{tab:results-exp2} summarizes the results of the four state of the art concept learners and \celsystem{}-RS on the 7 learning problems.
\celsystem{}-RS achieves significantly better F1 scores than \celoe{} and \drill{}.\footnote{We use a Wilcoxon signed-rank test with $\alpha=0.05$.} 
A deeper analysis of the behavior of \celoe{} and \drill{} shows that these approaches already need more than the provided 10 minutes to execute $\refine(\top)$, i.e., the first step at the beginning of their search. 
\evolearner{} and \nces{} did not give any results on the large knowledge bases of this experiment.\footnote{We worked together with the authors of \evolearner{} and \nces{} but couldn't find a solution before the submission deadline.}

\celsystem{}-RS provides significantly better F1-scores than \celoe{} and \drill{} for all 7 learning problems when relying on balanced accuracy or the F1 measure during the search. Consequently, it is also significantly better than a baseline that would always return the concept $\top$.
The usage of accuracy leads to mixed results and seems to mislead the search when the learning problem has only a small number of positive examples.\footnote{The full set of results of \celsystem{}-RS can be found in our Github project.}

\subsubsection{Experiment II}

\begin{table}[tb]
\caption{Survey results per question for the two learning problems (in \%, $\textcolor{niceblue}{\blacksquare}$ Yes (Y), $\textcolor{nicered}{\blacksquare}$ No (N)). The MV row shows the summary of the answers as a majority vote and the Exp row the expected results.}
\includegraphics[width=\linewidth]{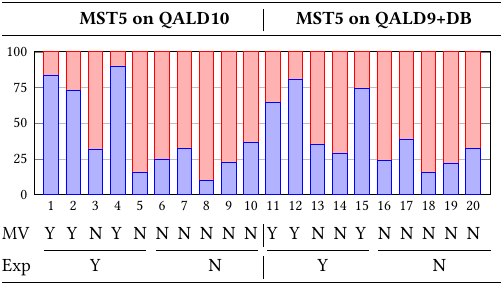}
\centering

\label{tab:survey-results}
\vspace{-3mm}
\end{table}

\surveyParticipants{} people participated in our survey, answering at least 1 question.
Table~\ref{tab:survey-results} shows the survey results. For 16 out of 20 questions (i.e., 80\%), the majority of the participants was able to correctly decide whether the described QA system would be able to answer the given question.
For all questions, except question 17, the answers of the volunteers are significantly different to those of a random guesser.\footnote{Significant = a random guesser has a probability below 5\% to create the answer set.}
A closer look at the questions that were not classified correctly by the majority of the participants revealed two patterns of errors.
First, the verbalization generated by ChatGPT was not always exactly explaining the given concept. For example, in the case of \mst{} on QALD10, while one of the properties has the label "manifestation of" ChatGPT translated it into "something that has a type or form", leading to people ignoring triples with the "manifestation of" property in the provided data.
Second, in the case of MST5 on QALD9+DB, ChatGPT (or the participants) relied on background knowledge that was not part of \knowledgebase{}. This led to a misunderstanding $\neg\text{agent}$ that was incorrectly understood as a non-person by ChatGPT.
This underlines the importance of a well-defined ontology as basis for the concept learning and, hence, for our approach.

\section{Discussion}
The results of Experiment I show that due to \celsystem{}'s scalability, it outperforms the other four approaches on knowledge bases containing tens of millions of triples. The importance of the scalability is further underlined when taking into account the sizes of real-world knowledge graphs like DBpedia~\cite{Auer2007DBpedia} or Wikidata~\cite{Vrandecic2014Wikidata}.

In Experiment I, \celsystem{} achieves F1 scores that are higher than the performance of the concept $\top$ would be. This suggests that our approach is able to provide meaningful explanations.
This is supported by the results of Experiment II which suggest that humans can accurately predict the behavior of the benchmarked system based on these explanations. However, future work will have to take the two identified sources of errors into account.





\section{Conclusion}

We proposed an approach for generating explainations for benchmarking results.
Our approach relies on the transformation of the benchmarking dataset into a knowledge base and on concept learning to find a concept that separates cases in which the benchmarked system performed good from those in which it performed subpar.
We also proposed a new, scalable concept learning algorithm named \celsystem{}, which uses the monotonicity of subset inclusion to prune its search tree.
Our evaluation used benchmark datasets and real-world benchmarking results from the knowledge-graph-based Question Answering domain. 
Our evaluation results show that \celsystem{} significantly outperforms state-of-the-art concept learners on the knowledge bases created by our approach due to its scalability.
A survey including the answers of \surveyParticipants{} participants showed that in 80\% of the cases the majority of participants were able to understand the explanations our approach generates for the evaluation results of QA systems.

Our future work is threefold.
First, we plan to apply our generic approach to other application areas. 
Second, we want to further improve existing concept learners. 
Third, a large scale experiment is needed to ensure that the generated explanations cannot only be understood be the explainee, but also support them in improving the benchmarked system over time.

\begin{acks}
This work has been supported by the Ministry of Culture and Science of North Rhine-Westphalia (MKW NRW) within the projects SAIL (NW21-059D) and WHALE (LFN 1-04, under the Lamarr Fellow Network programme), and the European Union’s Horizon Europe research and innovation programme in the project ENEXA (No. 101070305).
\end{acks}

\bibliographystyle{ACM-Reference-Format}
\bibliography{main}

\end{document}